\documentclass[letterpaper, 10 pt, journal, twoside]{support/IEEEtran}
\usepackage[normalem]{ulem}
\usepackage{soul}
\usepackage{xcolor}
\usepackage{etoolbox}
\usepackage{amsmath}
\usepackage{amsfonts}
\usepackage{setspace}
\usepackage[space, compress, sort]{cite}

\usepackage{amsthm}
\usepackage{amssymb,stmaryrd}
\usepackage[ruled,vlined]{algorithm2e}
\usepackage{mathtools}
\usepackage{tabularx}
\usepackage{graphicx}
\usepackage{subfigure}
\usepackage{enumerate}
\usepackage{float}
\usepackage{url}
\usepackage{verbatim}
\usepackage{booktabs} 
\usepackage{multirow}
\usepackage[linkcolor=black,citecolor=black,urlcolor=black,colorlinks=true]{hyperref}
\usepackage{bm}
\graphicspath{{figures/}}
\IEEEoverridecommandlockouts

\newcommand{\delete}[1]{{\bgroup\markoverwith{\textcolor{red}{\rule[0.5ex]{2pt}{0.4pt}}}\ULon{#1}}}
\newcommand{\deletefig}[1]{{\bgroup\markoverwith{\textcolor{red}{\rule[2.5ex]{2pt}{2.0pt}}}\ULon{#1}}}

\setstcolor{red}

\makeatletter \renewcommand{\@thesubfigure}{\thesubfigure \space}

\newcommand{\tabincell}[2]{\begin{tabular}{@{}#1@{}}#2\end{tabular}}

\begin{document}
	\title{External Forces Resilient Safe Motion
		Planning  \\ for Quadrotor}\label{key}
	\author{ Yuwei Wu$^{1,2}$, Ziming Ding$^{1}$, Chao Xu$^{1}$ and Fei Gao$^{1}$
	\thanks{Manuscript received: March 17, 2021; Revised: June 25, 2021; Accepted: August 18, 2021. This paper was recommended for publication by Editor Hanna Kurniawati upon evaluation of the Associate Editor and Reviewers' comments. This work was supported by National Natural Science Foundation of China under Grant 62088101 and Grant 62003299. (\textit{Corresponding author: Fei Gao, Chao Xu}) }
	\thanks{$^{1}$State Key Laboratory of Industrial Control Technology, Institute of Cyber-Systems and Control, Zhejiang University, Hangzhou 310027, China, and Huzhou Institute of Zhejiang University, Huzhou 313000, China.
	 }
	\thanks{$^{2}$Department of Electrical and Systems Engineering, University of Pennsylvania, Philadelphia, PA 19104 USA}
	\thanks{E-mail: {\tt\small yuweiwu@seas.upenn.edu, \{zm\_ding, cxu, fgaoaa\}@zju.edu.cn }}
	\thanks{Digital Object Identifier (DOI): see top of this page.}}

	\maketitle
    \begin{abstract}

        Adaptive autonomous navigation with no prior knowledge of extraneous disturbance is of great significance for quadrotors in a complex and unknown environment. The mainstream approach that considers external disturbance is to implement disturbance-rejected control and path tracking. However, the robust control to compensate for tracking deviations is not well-considered regarding energy consumption, and even the reference path will become risky and intractable with disturbance. As recent external forces estimation advances, it is possible to incorporate a real-time force estimator to develop more robust and safe planning frameworks. This paper proposes a systematic (re)planning framework that can resiliently generate safe trajectories under volatile conditions. Firstly, a front-end kinodynamic path is searched with force-biased motion primitives. Then we develop a nonlinear model predictive control (NMPC) as a local planner with Hamilton-Jacobi (HJ) forward reachability analysis for error dynamics caused by external forces. It guarantees collision avoidance by constraining the ellipsoid of the quadrotor body expanded with the forward reachable sets (FRSs) within safe convex polytopes. Our method is validated in simulations and real-world experiments with different sources of external forces. 
    \end{abstract}
	
	\begin{IEEEkeywords}
	  Motion and Path Planning; Aerial Systems: Applications; Collision Avoidance
	\end{IEEEkeywords}
	
	\IEEEpeerreviewmaketitle
		\vspace{-0.3cm}
	\section{Introduction}
	\label{sec:introduction}
	Safe trajectory generation under real-world conditions where quadrotors are exposed to external disturbance and other sources of uncertainty has been a crucial concern, especially in complex and dynamic environments~\cite{6630661,8429263,1657388}. External forces caused by air turbulence, loaded objects, or other uncertain disturbance induce a considerable effect on the quadrotor system and challenge planning stability. 

\begin{figure}[!t]
	\centering
	\begin{center}	
		\label{fig:real12}
		{\includegraphics[width=1\columnwidth]{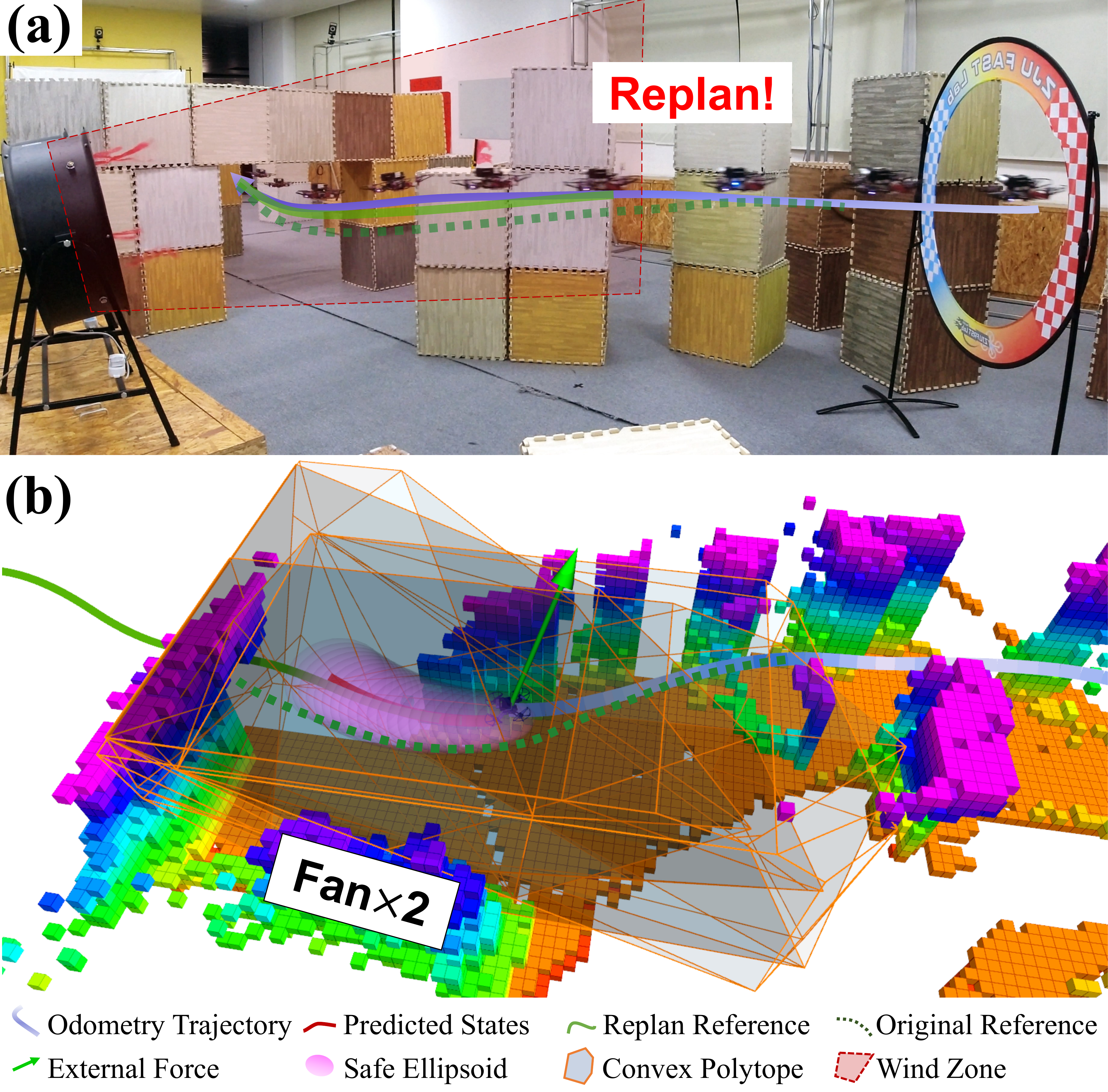}}
	\end{center}
	\vspace{-0.3cm}	
	\caption{\label{fig:fig1} An indoor flight when the quadrotor enters a wind zone. (a) The wind pushes the quadrotor to obstacles while our planner replans a new feasible reference (green dotted) to adapt. (b) Visualizations. Video is at: https://youtu.be/nSKbzAM0v18 \label{fig:indoor_experiments}}
	\vspace{-0.6cm}
\end{figure}
Existing attempts mainly focus on robust control with unpredictable disturbance to prevent the quadrotor from deviations to its desired trajectories. Disturbance-aware controllers for quadrotors like robust MPC~\cite{KAMEL20173463} and tube-based MPC~\cite{7989693} aim to follow a predefined reference path with robustness guarantees and try to compensate for the influence of the disturbance. However, these methods either may fail to handle tremendous forces because of the controller's limitations or generate too conservative trajectories because they consider all sources of uncertainty. Another work \cite{ji2020cmpcc} addresses this issue by adding a receding horizon, corridor-based, local adaptive tracking layer between the planner and the controller. When the quadrotor encounters an external force, this method directly replans a new trajectory on the current state without any estimation of this force. Due to its lack of modeling on the external force, it can not handle sustained forces and fails to solve a feasible path. Other works partly solve this problem by planning under known constant wind \cite{1657388} or with prior information of a force field \cite{7487414}. These methods partially simplify the environments and need reliable predictions of the external force. As a result, it is impractical in real-world situations with dynamic and unexpected external disturbances.

In this work, we propose a systematic framework to achieve robust local planning, which accounts for the influence of extreme external disturbance \footnote{The source code is available at \url{https://github.com/ZJU-FAST-Lab/forces_resilient_planner}}. 
The presented method searches a dynamically feasible path by kinodynamic hybrid-state A* \cite{boyu2019ral} as a rough reference with added nominal force estimated by the method in \cite{ding2020vidfusion} to generate motion primitives. Then we formulate an optimal control problem (OCP) for trajectory generation based on the reference path, which encodes the external force with its uncertainty. This OCP can be solved by a corridor-based NMPC that exploits the full dynamics and geometrical shape of a quadrotor. We extend the ellipsoidal approximation of FRSs\cite{8968126} which are the state sets of error dynamics in a planning horizon. Therefore, we can constrain the discrete-time FRSs in a safe flight corridor for collision avoidance. The real-world experiments illustrate the robustness of our method, and an indoor flight is shown in Fig. \ref{fig:fig1}.

Our contributions can be summarized as:
\begin{itemize}
	% the front-end
	\item An efficient front-end kinodynamic path searching and collision checking method with external forces consideration. 
	
	% the back-end optimizations
	\item An online trajectory optimization algorithm that utilizes nonlinear model predictive control with safe ellipsoid boundaries constrained in a convex flight corridor to enforce reliable obstacle avoidance.
	
	% the overall systems
	\item A (re)planning framework for quadrotor that integrates the VID-Fusion external force estimator \cite{ding2020vidfusion}. The onboard experiments demonstrate the performance of our system in volatile real-world environments.
\end{itemize}
	
	\section{Related Work}	
	\label{sec:related_works}
	\subsection{Modeling of External Forces}

% external forces expression: wind field or the estimator/observer
Accurate estimation and prediction of external forces assist robust control\cite{6907146} and offset-free trajectory tracking for a quadrotor. Some works attempt to present this special external force by modeling the disturbance as a predicted spatiotemporally varying wind field \cite{7487414} or a Gaussian mixture field~\cite{7989731}. For more generic conditions with unknown types of external forces, different force estimators are employed. Extended Kalman Filter (EKF) is applied in \cite{KAMEL20173463} to estimate the external disturbance while utilizing the same model used in the inner-loop attitude dynamics to decrease the tracking errors. Other approaches like nonlinear observer~\cite{6878116}, momentum-based estimator \cite{6907146}, and optimization-based methods \cite{ding2020vidfusion, 8421746} are also widely used for visual-inertial odometry (VIO) systems and force estimation. 

To further include the external force into a quadrotor system, Kamel et al.\cite{KAMEL20173463} directly add an estimated force to the dynamic model. Other methods model the disturbance as a zero-mean, bounded disturbance \cite{8968126} with handcrafted bound modification when the drone encounters sudden winds, or as Gaussian uncertainty distributions propagated with 3$-\sigma$ confidence ellipses \cite{8202163}. These approaches can guarantee the safety and robustness of a quadrotor system but lack the flexibility to handle excessive disturbance. With an external force estimated in \cite{ding2020vidfusion}, we can apply a tighter bounded noise as the uncertainty of external force into the model.

\subsection{Robust Planning With Disturbance}

% the control and tracking with disturbance
Researchers have focused on OCP of tracking a given reference path with robustness consideration to handle uncertainties and disturbances. MPC is an efficient tool to generate optimal local trajectories while encodes uncertainty and obstacle avoidance. Although linear MPC demonstrates its excellent performance on trajectory generation and navigation \cite{ji2020cmpcc,BANGURA201411773}, nonlinear MPC of full dynamics is better regarding disturbance rejection, especially for aggressive trajectory tracking\cite{KAMEL20173463, 8202163}. 

The effects of external forces require a quadrotor system to apply higher-level strategies to resist disturbances and plan a safe and energy-efficient path. Among all attempts for robust path planning, Singh et al. \cite{7989693} generate a conservative trajectory subject to bounded disturbances by pre-computing a globally valid invariant tube, while others \cite{6630661,7989731} conduct global planning assuming that a quadrotor is flying in a region of known time-varying winds. However, such methods rely on the accuracy of disturbance predictions and modeling, which are not practicable for real-world planning. To solve these limitations, we incorporate kinodynamic path searching and NMPC local planning to generate safe trajectories with real-time estimated force. In terms of obstacle avoidance and free space representation, the common method represents the obstacles by flexible parameterization \cite{8968126} and incorporates cost functions with the distance to the nearest obstacle.  Another representation of free space in an environment generates a safe flight corridor of several convex polytopes ~\cite{gao2020teach, 7839930, zhong2020generating} to constrain the trajectory. Our method also exploits such representation\cite{zhong2020generating} to generate large convex polytopes directly on an occupancy map.

  \section{Planning With External Forces}
  \label{sec:planning with external forces}

\subsection{Quadrotor Dynamic Model}
We consider the dynamic model with accurate control of Euler angles, which assumes that its rate can accurately track the desired command. The system state is $  \bm{x} = [ \bm{p}, \bm{v}, \phi, \theta, \psi ]^{T}  \! \in \mathcal{X} \! \subset \!  \mathbb{R}^{n_x} $, where $\bm{p} = [p_{x}, p_{y}, p_{z}]^{T} ,  \bm{v} = [v_{x}, v_{y}, v_{z}]^{T}  \! \in \! \mathbb{R}^{3} $ denote the position and velocity of the quadrotor. $\phi$, $\theta$, $\psi \in \mathbb{R} $ are the roll, pitch and yaw angles. The control input is $\bm{u} \! = \!  [\dot{\phi}_{c},\dot{\theta}_{c},  \dot{\psi}_{c},  T_{c}  ]^{T} \!  \in \!  \mathcal{U}  \! \subset\!\mathbb{R}^{n_u} $ in which $\dot{\phi}_{c}, \dot{\theta}_{c},  \dot{\psi}_{c} \in \mathbb{R}$ are the command rates of the Euler angles, $T_{c} \in \mathbb{R}$ is thrust command of the quadrotor in body frame. We provide a full-body nonlinear dynamic model as follows:
\vspace{-0.1cm}
\begin{subequations}
	\begin{align}
	\dot{\bm{p}} &=\bm{v},  \\
	\dot{\bm{v}} &= \frac{1}{m}( \bm{R}\begin{bmatrix}0\\ 0\\  T_{c}\end{bmatrix} - \bm{F}_{drag} + \bm{F}_{ext}  )- \begin{bmatrix}0\\ 0\\ g\end{bmatrix}  \label{Za},  \\
	\dot{\phi} &= \dot{\phi}_{c},  \\
	\dot{ \theta} &= \dot{\theta}_{c},   \\
	\dot{\psi} &= \dot{\psi}_{c},
	\end{align}
\end{subequations}		
where $\bm{R} \! \in \! \mathbb{R}^{3 \times 3}$ is the rotation matrix parameterized by the Euler angles, $g \! \in \! \mathbb{R}$  is the magnitude of gravitational acceleration, $m \! \in \! \mathbb{R}$  is the mass of the quadrotor. We can represent the nonlinear dynamics as $\dot{\bm{x}} = f(\bm{x}(t),\bm{u}(t), \bm{F}_{ext}(t))$. Components of the rotor drag and drag like effects are complicated to model and correlated with quadrotor's velocity, we consider it in the system to compensate its interference to disturbance bound\cite{8387437}. To further simplify its aerodynamics, we apply a first-order drag model by adding the drag force $\bm{F}_{drag}$ in the x$-$y body frame. We define $\bm{K}_{drag} := diag\{k_d, k_d, 0 \} $, where $k_d$ is the drag coefficient constant, then the drag force in the world frame is
\begin{equation}
\label{eq:drag}
\bm{F}_{drag}  = \bm{R}  \bm{K}_{drag} \bm{R}^{T} \bm{v}. \\
\end{equation}

Instead of predicting the external forces or modeling the field of special force like winds, we generally obtain a real-time force estimation and then adjust the planning strategy when the force surpasses its bound. The disturbance of external forces can be expressed as a nominal value with an additive bounded noise, which makes it possible to design an NMPC that satisfies the constraints under bounded disturbance. The nominal force $\bm{b}_{ext}$ can be treated as a constant value calculated in~\cite{ding2020vidfusion} for a sufficiently short duration. The external force is defined as $\bm{F}_{ext} :=  \bm{b}_{ext}  + \bm{w}_{ext} $, where the bounded noise $ \bm{w}_{ext} \in \mathbb{W} = \{ \bm{w} \in \mathbb{R}^{n_{w}}  :||\bm{w}||_{\infty } \leq w_m \} $, $w_m$ is the maximum bound.
		
\subsection{Collision Avoidance Constraints}

\subsubsection{Error Dynamics}

We construct a linearized system for closed-loop dynamics of error state $\bm{e}(t) =  \bm{x}(t)- \bm{\underline{x}}(t)$ around nominal states $\bm{\underline{x}}(t) $, inputs $\bm{\underline{u}}(t)$ and nominal external force $\bm{b}_{ext}$. Assuming that the linearization error can be neglected around nominal states with the condition that the feedback control policy can be expressed as  $\bm{u}(t) = \bm{K}(t)\bm{e}(t) + \bm{\underline{u}}(t)$, $\bm{K}(t)$ is the feedback gain. We can get the error dynamics as
\begin{equation}
\label{eq:error}
\bm{\dot{e}}(t) = \bm{\Gamma}(t)\bm{e}(t) + \bm{D}(t) \bm{F}_{ext}(t),
\end{equation}
where $ \bm{\Gamma}(t) =  \bm{A}(t) + \bm{B}(t)  \bm{K}(t)$ , $ \bm{A}(t) := \left.\partial f/\partial \bm{x} \right|_{(\underline{\bm{x}},\underline{\bm{u}}, \bm{b}_{ext} )}$, $\bm{B}(t) := \left.\partial f/\partial \bm{u} \right|_{(\underline{\bm{x}},\underline{\bm{u}}, \bm{b}_{ext} )} $, $ \bm{D}(t):= \left.\partial f / \partial\bm{F}_{ext} \right|_{(\underline{\bm{x}},\underline{\bm{u}}, \bm{b}_{ext} )}$.
\vspace{0.1cm}

In order to propagate the uncertainty of external forces on each nominal state, we apply the analytic solution based on Hamilton-Jacobi (HJ) reachability analysis that quantitatively expresses the error FRS $ \mathcal{E}(t)$ as approximate ellipsoids\cite{8968126}. With a more precious estimation of the external forces by VID-Fusion\cite{ding2020vidfusion}, a less conservative error FRSs could be obtained by employing a tighter variance bound.

\subsubsection{Propagation of Safe Ellipsoid Boundary}

To further solve the problem, we discrete the dynamics with a sampling time $t_s$ over $N$ time steps,  as $ \bm{x}^{k+1} = f_d( \bm{x}^{k}, \bm{u}^{k}, \bm{F}_{ext}^{k} )$. For convenience,
we denote the discrete-time state as $\bm{x}^{k} =  [ \bm{p}^{k}, \bm{v}^{k}, \phi^{k}, \theta^{k}, \psi^{k} ]^{T} $ at stage $k$, $\forall k \in \{ 0,1,\ldots, N-1 \}$. The rotation matrix is $\bm{R}^{k} $, the error FRS is $ \mathcal{E}^k$, and the external force is denoted as $\bm{F}_{ext}^{k} = \bm{b}_{ext} + \bm{w}_{ext}^{k}$.

We inflate the geometrical shape of a quadrotor to ensure the safety of our generated trajectory. The quadrotor is modeled as an ellipsoid with a radius $r$ and a height $h$. Therefore, we can represent ego geometrical ellipsoid at stage $k$ as
	\begin{equation}
	\label{eq:ellipsoid}
	\xi (\bm{p}^{k}, \bm{Q}_{ego}^{k}) := \{ \bm{p} \in  \mathbb{R}^{3} : (\bm{p}-\bm{p}^{k})^{T}      (\bm{Q}_{ego}^{k})^{-1} (\bm{p}-\bm{p}^{k}) \leq 1 \}, \\
	\end{equation} % Q is symmetric and positive definite
	which is centered on quadrotor's position $\bm{p}^{k}$ with a shape matrix as
	\begin{equation}
	\label{eq:ego}
	\bm{Q}_{ego}^{k} = \bm{R}^{k}  {\rm diag}\{r^2, r^2, h^2 \} (\bm{R}^{k})^{T} \in  \mathbb{S}^{3}_{+}.
	\end{equation} 
	
\begin{figure}[!th]  
		\centering
		{\includegraphics[width=0.85\columnwidth]{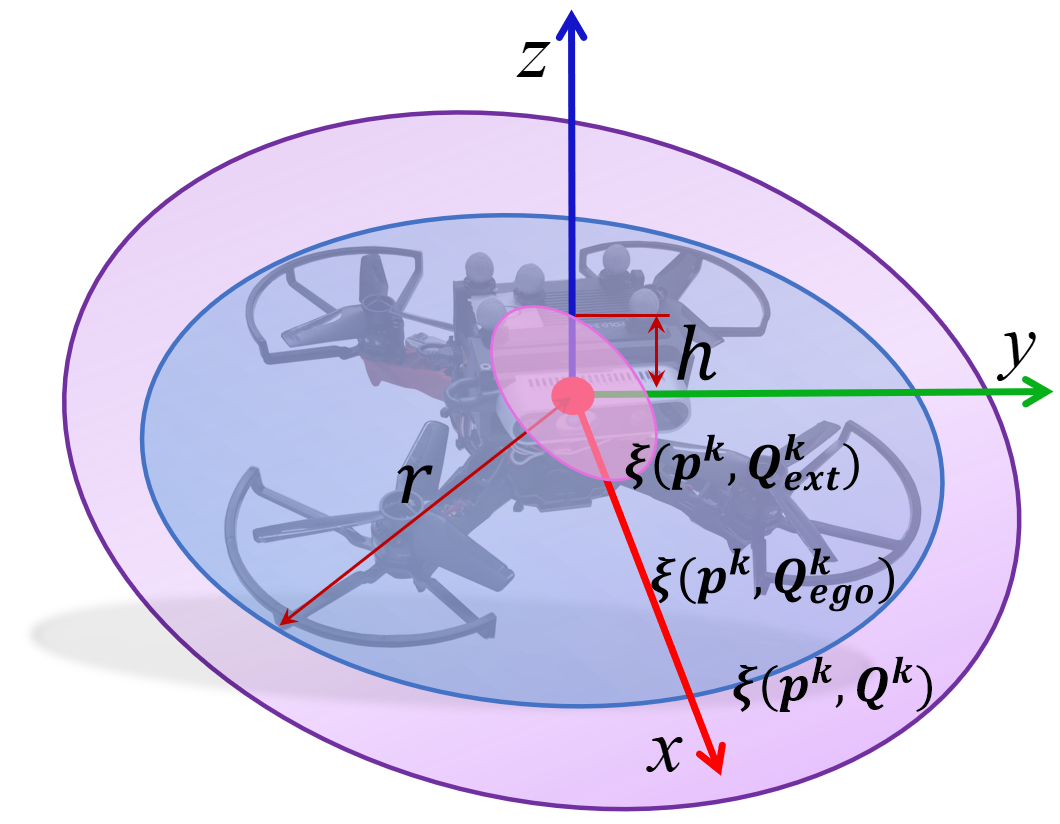}}
		\caption{The inner ellipsoid (in pink) is the ellipsoidal approximation of the error forward reachable set, the middle one (in blue) is the geometrical modeling of the quadrotor. The outer ellipsoid (in purple) is the approximation of ellipsoidal sum. \label{fig:ellpsoid}}
		\vspace{-0.3cm}
	\end{figure}

	As shown in Fig. \ref{fig:ellpsoid}, the safe ellipsoid boundary $\xi (\bm{p}^{k}, \bm{Q}^{k})$ is an outer ellipsoid that includes the shape of a quadrotor $\xi (\bm{p}^{k}, \bm{Q}_{ego}^{k} )$ and ellipsoidal approximation of error FRS $\mathcal{E}^k = \xi (\bm{p}^{k}, \bm{Q}_{ext}^{k} )$ , where $\bm{Q}^{k}, \bm{Q}_{ext}^{k} \in \mathbb{S}^{3}_{+} $ are the shape matrices accordingly. Given two ellipsoids centered on the same point, the optimal outer ellipsoidal approximation which is guaranteed to contain the Minkowski sum \cite{10.1007/978-1-4612-4120-1_12} of these ellipsoids is denoted by
	
	\begin{equation}
	\label{eq:q_example}
	\bm{Q}_{1}  \boxplus  \bm{Q}_{2}  =  (1+\beta) \bm{Q}_{1} +(1+\frac{1}{\beta}) \bm{Q}_{2},\\
	\end{equation}
	where $\beta = \sqrt{\frac{tr(\bm{Q}_{2})}{tr(\bm{Q}_{1})}}$, $\bm{Q}_{1}, \bm{Q}_{2} \in  \mathbb{S}^{3}_{+}$ are the shape matrices of ellipsoids, $\mathbb{S}^{3}_{+}$ denotes the real symmetric positive definite matrices.
	\vspace{-0.3cm}  

	Therefore, the shape matrix $\bm{Q}^{k} \in  \mathbb{S}^{3}_{+}  $ of the safe ellipsoid boundary is obtained by  $\bm{Q}^{k} =  \bm{Q}_{ego}^{k}  \boxplus  \bm{Q}_{ext}^{k} $. Because we use the discrete representation of the model, we propagate $\bm{Q}_{ext}^{k}$ in each stage with current error initial shape matrix $ \bm{Q}_{0}^{k}$ and system matrix $\bm{\Gamma}^k$ of quadrotor rather than using the same initial state during a sampling time $t_s$. The max outer ellipsoid propagation law of error FRS can be computed by
	\begin{equation}
	\label{eq:qext}
	\bm{Q}_{ext}^{k} \! \! =  \! \left [  {\rm exp}  (\bm{\Gamma}^k t_s)  ( \bm{Q}_{0}^{k} \boxplus  \bm{Q}_{d}^{k}) \  {\rm exp} (\bm{\Gamma}^{k,T} t_s)  \right ]_{3\times3}.  \\
	\end{equation}

	Note that we only concern the uncertainty of quadrotor's position in the total error ellipsoid for further obstacle avoidance, so we directly block the shape matrix to the first 3-dimension matrix,  where $ \left [\cdot \right ]_{3\times3} $ denotes the first $3\!\times\!3$ block of the matrix. The maximal ellipsoidal reachable set $ \bm{Q}_{d}^{k}$ of error dynamics is individually computed by~\cite{8968126} with the converted discrete-time model. The error initial shape matrix $\bm{Q}_{0}^{k+1}$ is updated with the outer approximation as $\bm{Q}_{0}^{k+1} = \bm{Q}_{0}^{k} \boxplus \bm{Q}_{d}^{k}$.

	\subsubsection{Safe Corridor Constraints}	
\begin{figure}[thpb]  
	\vspace{-0.55cm}  
	\centering
	{\includegraphics[width=0.99\columnwidth]{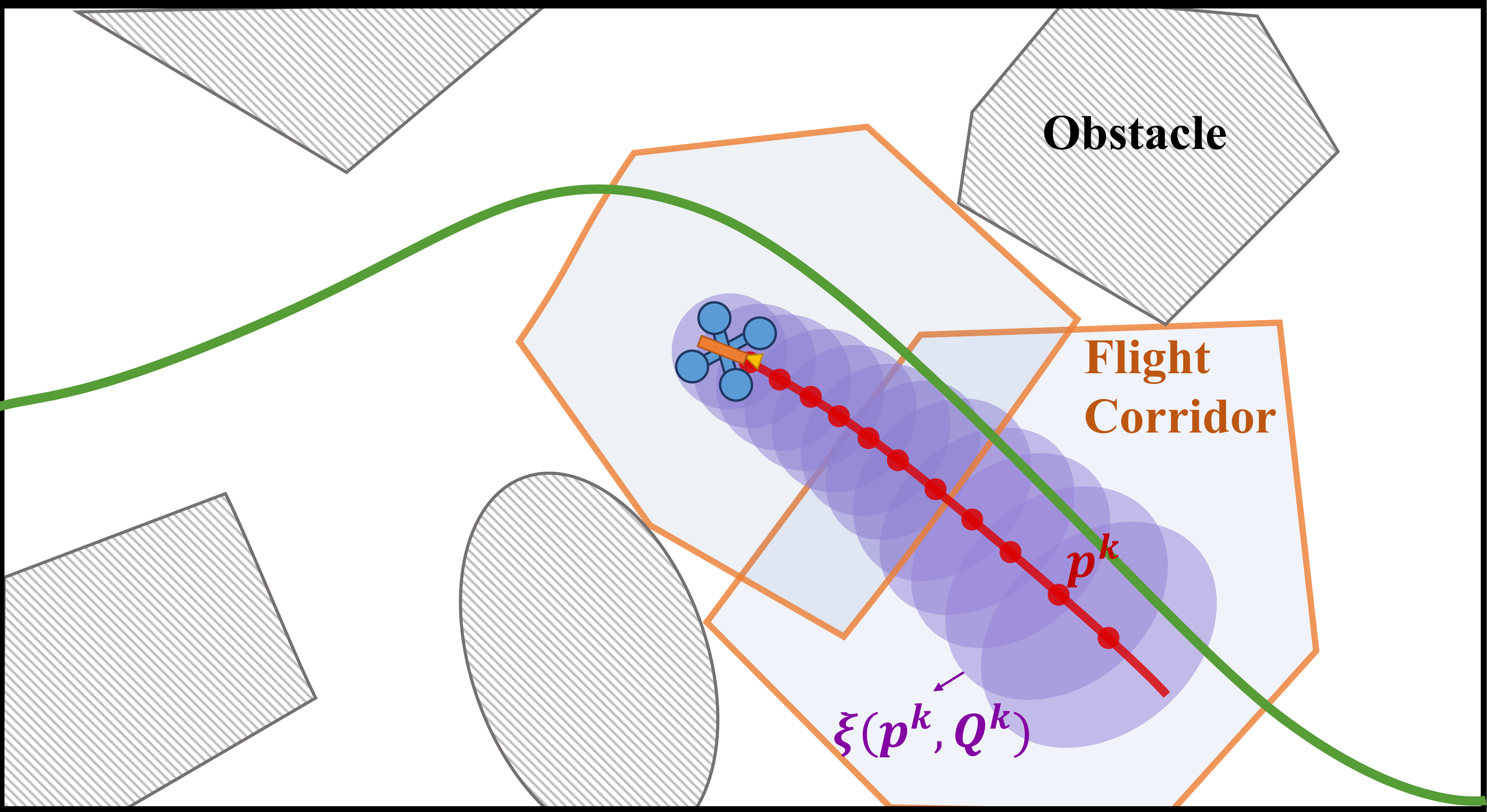}}
	\caption{\label{fig:const} Safety constraints by restricting the safe ellipsoids within a fight corridor. In the planning horizon, the safe ellipsoid boundary at stage k is $\xi (\bm{p}^{k}, \bm{Q}^{k})$, as shown in Fig. \ref{fig:ellpsoid}.}
	\vspace{-0.5cm}
    \end{figure}

	With the propagated safe ellipsoid boundary in each stage, we can encode collision avoidance constraints by restricting the ellipsoid in a convex collision-free corridor. To ensure sufficient freedom for NMPC to get a better and refined trajectory, we use the method in~\cite{zhong2020generating} to rapidly generate a series of convex polytopes to cover reference waypoints in the current planning horizon, as shown in Fig. \ref{fig:const}. The convex polyhedron can be represented as linear constraints with $ \bm{A}_{i} \in \mathbb{R}^{m \times 3}$ and $ \bm{b}_{i} \in \mathbb{R}^{m}$, $ \forall i \in  \{ 1,2, \dots, m \} $, pre-assigned with the reference waypoints. Inspired by the computation of the maximum volume inscribed ellipsoid in a polytope\cite{boyd2004convex}, we have
	\begin{equation}
	 \{ \bm{p} :  \bm{p} \in \xi (\bm{p}^{k}, \bm{Q}^{k} ) \} \in \{ \bm{p}: \bm{A}_{i}\bm{p}\leq \bm{b}_{i}  \}.
	\end{equation}
	It is equal to solve the problem
    \begin{subequations}
		\begin{align}
		&h(\bm{p}^{k},\bm{Q}^{k}) = \max \bm{A}_{i}\bm{p}\leq \bm{b}_{i}, \\
		\rm{s.t.} &\ (\bm{p}-\bm{p}^{k})^{T} ( \bm{Q}^{k})^{-1}(\bm{p}- \bm{p}^{k}) \leq 1.
		\end{align}
		\setlength\belowdisplayskip{0.1pt}
	\end{subequations}
	This optimization problem can be easily solved analytically, so that the collision avoidance constraint can be obtained by
	\begin{equation}
	h(\bm{p}^{k},\bm{Q}^{k}) =  \left \| (\bm{Q}^{k})^{\frac{1}{2}} \bm{A}_{i}^{T}  \right \|+ \bm{A}_{i}\bm{p}^{k} \!\leq \bm{b}_{i}.
	\end{equation}
	\subsection{Objective Functions}
    For robust reference tracking and yaw control, while reducing the control efforts, we penalize the cost with the following three terms.		
	\subsubsection{Navigation Cost}
    In each stage, we minimize the tracking deviation from the reference position to the predicted one. Other references as velocity and acceleration are relaxed during the optimization. A yaw angle sequence is pre-computed according to the orientation of reference velocity of the front-end path. We define the stage cost as
	\begin{equation}
	J_{\bm{x}}^{k}( \bm{x}^{k})= \left \| \bm{p}^{k} - \bm{p}^{k}_{ref} \right \|_{\bm{l}_{p}}  +  \left \| \psi^{k} - \psi^{k}_{ref} \right \|_{\bm{l}_{\psi}},
	\end{equation}
    where $ \left \|\cdot \right \|_{l_{p}} $ and $ \left \| \cdot \right \|_{l_{\psi}}  $ are
    weighted squared norm induced by the $\bm{l}_{p}$ and $\bm{l}_{\psi}$ matrices. The terminal cost is
    
    \begin{equation}
    J^{N}( \bm{x}^{N}) = \left \| \bm{p}^{N} - \bm{p}^{N}_{ref} \right \|_{\bm{l}^{N}_{p}}  +  \left \| \psi^{N} - \psi ^{N}_{ref} \right \|_{\bm{l}^{N}_{\psi}},
    \end{equation}  
    where ${\bm{l}^{N}_{p}}$ and ${\bm{l}^{N}_{\psi}}$ are the terminal weighting matrices. When the remained reference state is beyond its planning horizon, a velocity penalty term $ \left \| \bm{v}^{N} \right \|_{\bm{l}^{N}_{v}}$ is added in the final stage to approach the terminal condition of rest.

	\subsubsection{Control Input Cost} 
	In the aspect of control efforts, the penalty of the control input should be added as follows:
	\begin{equation}
	J_{\bm{u}}^{k}(\bm{u}^{k})=\left \| \bm{u}^{k} \right \|_{\bm{l}^{k}_{u}},
	\end{equation}
	$\bm{l}^{k}_{u}$ is the weighting matrix of control input cost.
	
	\subsubsection{Control Input Rate Cost}  
	For the smoothness of the control input, we also penalize the variations between current and previous inputs for avoiding oscillatory trajectories, as
	\begin{equation}
	J_{\Delta \bm{u}}^{k}(\bm{u}^{k+1},\bm{u}^{k})= \left \|\bm{u}^{k+1}-\bm{u}^{k}\right \|_{\bm{l}^{k}_{\Delta \bm{u}}},
	\end{equation}
	$\bm{l}^{k}_{\Delta \bm{u}}$ is the weighting matrix of the smoothness cost.
	
	\subsection{External Forces Resilient NMPC}
	Finally, we can formulate the receding horizon optimization problem with nominal external force as
	\begin{subequations}
		\begin{align}
		\min_{x, u} J^{N}(\bm{x}^{N}) +\sum_{k=0}^{N-1}J_{\bm{x}}^{k}(\bm{x}^{k}) + J_{\bm{u}}^{k}(\bm{u}^{k})+J_{\Delta \bm{u}}^{k}(\bm{u}^{k+1}, \bm{u}^{k}), \label{eq :totalobj}
		\end{align}
		\setlength\abovedisplayskip{0.1pt}
		\begin{align}
		{\rm s.t.}  & \quad \bm{x}^{k+1} =f_d(\bm{x}^{k}, \bm{u}^{k}, \bm{b}_{ext}),  
		\quad  \label{eq :transit} \quad \quad \quad\\		
		& \quad \bm{x}^{0}=\bm{x}_{0},  \\
		&  \quad h(\bm{p}^{k},\bm{Q}^{k}) \leq \bm{b}_{i}, \quad  \\
		&  \quad \bm{Q}^{k} =  \bm{Q}_{ego}^{k}  \boxplus  \bm{Q}_{ext}^{k}, \label{eq:Qk} \\
		&  \quad \bm{u}^{k} \in \mathcal{U},  \quad \bm{x}^{k}  \in \mathcal{X}  \label{eq:input_state},
		\end{align}
	\end{subequations}
    where $\bm{x_{0}} \in \mathcal{X}$ is the initial state, $ \bm{Q}_{ext}^{k} $ and $\bm{Q}_{ego}^{k}$ in (\ref{eq:Qk}) are computed and propagated by equation (\ref{eq:ego}) - (\ref{eq:qext}). To reduce the problem's complexity, $\bm{Q}^{k}$ is pre-computed before each optimization using the last loop planned series of Euler angles. Upper and lower bounds for control inputs and states should also be considered in (\ref{eq:input_state}). 

	\subsection{Path Searching With Nominal Force}
	\label{subsec:Path Searching with Nominal Force}
	The reference collision-free path we mention above is generated based on a kinodynamic hybrid-state A* proposed in \cite{boyu2019ral}, which searches a kinodynamic feasible trajectory in an occupancy map that minimizes time duration and control cost. When expanding the motion primitives, we extend the simplified kinematic model with added nominal external force, which is better for computational efficiency compared with full nonlinear dynamics. This makes reference local path reliable under the specific external force instead of searching a global path which may collide if the system suffers such force. The front-end trajectory can be represented in three dimensions as time-parameterized piece-wise polynomials , ${\rm \Phi}(t) =  [ {\rm \Phi}_x(t), { \rm \Phi}_y(t), { \rm \Phi}_z(t)]^T $. The state is $\bm{s}(t) $$:= [{ \rm \Phi}(t)^T, \dot{{ \rm \Phi}}(t)^T]^T $, and the control input is $\bm{r}(t) =\ddot{{\rm \Phi}}(t)$. Hence, the state-space model is
	\begin{equation}
	\label{eq:kino_dynamic}
	\dot{\bm{s}} = \begin{bmatrix}
	0 & \rm{I_3} & 0\\ 
	0 & 0 & \rm{I_3}\\ 
	0 & 0 & 0
	\end{bmatrix} \bm{s} + \begin{bmatrix}0\\ 0\\ \rm{I_3} \end{bmatrix} ( \bm{r} + \frac{1}{m} \bm{b}_{ext} ). \\
	\end{equation}
    The collision-free path in the planning phase is naturally close to the obstacles and may become unsafe due to the variance of the external force. Therefore, rather than the exact path following, we only consider the feasible path as a rough reference for corridor generation and orientation guiding.

	\section{Implementation Details}
	\subsection{External Force Estimation}
	
	To ensure the robustness and accuracy of the estimation, a tightly-coupled Visual-Inertial-Dynamics estimator (VID-Fusion)~\cite{ding2020vidfusion} is used to optimize the pose and external force simultaneously via nonlinear optimization, which is suitable for our adaptive planning framework. 
	We exploit the estimated force in~\cite{ding2020vidfusion} as the sum of nominal external force and drag force, and then redefine the external force $\bm{F}_{ext}$ as a resultant force in the world frame except for rotor thrust, gravity, and drag force in our system.
	\subsection{System Overview of the (Re)Planning Framework}
	\subsubsection{External Forces Adaptation}
     We apply kinodynamic path searching\cite{boyu2019ral} with a constant nominal force to generate a front-end feasible path as reference. Because the update frequency of external force is much higher than replanning frequency, we introduce a noise bound to withstand the deviation of estimated external force in acceptable time duration. The back-end nonlinear MPC considers the error states as a series of ellipsoidal forward reachable sets, admitting external force to vary within this bound during the tracking phase. Therefore the followed waypoints and the optimized predicted states are guaranteed to be safe under such external forces.
	
	\subsubsection{Replanning Activation}
\begin{figure}[th]
	\centering
	{\includegraphics[width=1.0\columnwidth]{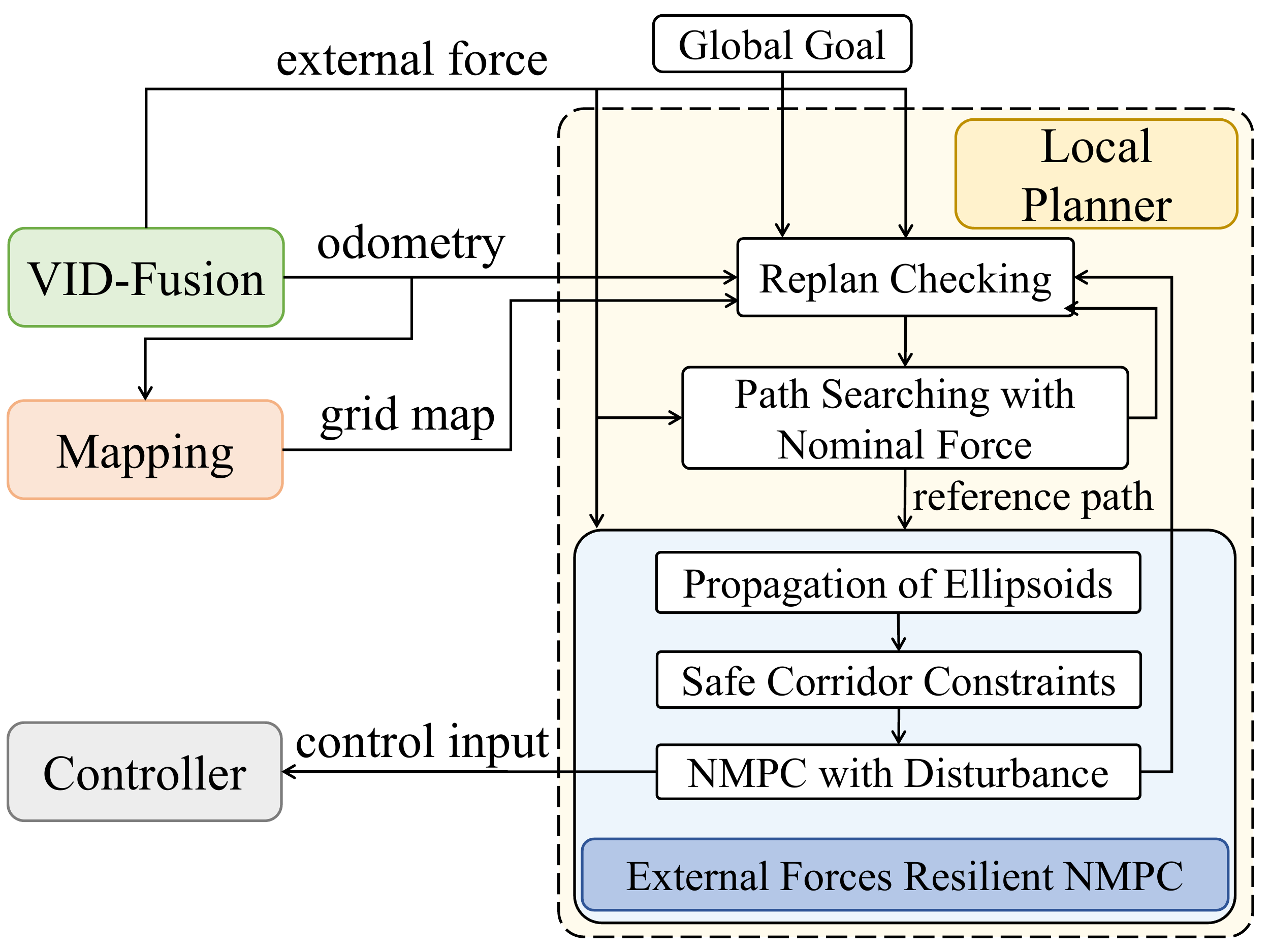}}
	\caption{The system overview diagram of the (re)planning framework. The VID-Fusion provides the odometry and estimated external force simultaneously, while our planning module checks this time-varying external force. If the force is within its allowed bound, the back-end NMPC is flexible to generate safe trajectories; otherwise, it triggers the replanning and reset reference path.}
	\label{fig:system}
	\vspace{-0.25cm}
     \end{figure}

	The replanning strategy is event-triggered under the feasible consideration of the external force and reference path, which combines the real-time external force both in front-end and back-end planning, as shown in Fig. \ref{fig:system}. When the variance of external force is upper the allowed bound, it is too fierce for NMPC to solve an acceptable solution on the current initial state. Then, the front-end reference needs to replan to fit with the current force. With the update of the occupancy map, if the global target or the reference path collides with obstacles, or the time-indexed reference path is hard to follow, then the replanning is triggered to generate a new path.

     \section{Results}
     \label{sec:results}
     \subsection{Setups}
     We use a code generation tool Forces Pro~\cite{FORCESPro} to solve our NMPC problem and directly use its generated C library for speed-up. We apply a receding horizon with $t_s = 50 \ \rm{ms} $ and total time steps $N=20$. The average solving time is around 5 ms on an Intel i7-10700 CPU computer for simulation tests and within 10 ms on our quadrotor platform. We define a mass-normalized external force $\bm{\mathcal{F}}^e = [\mathcal{F}_x^e, \mathcal{F}_y^e, \mathcal{F}_z^e]^T $ obtained by VID-Fusion. The noise bound of  $\bm{\mathcal{F}}^e$ is set as $0.5 \ \rm{m/s^2} $ according to our experiment data, to resist its allowable deviation during the planning phase.

\begin{figure}[!htbp]
	{\includegraphics[width=1\columnwidth]{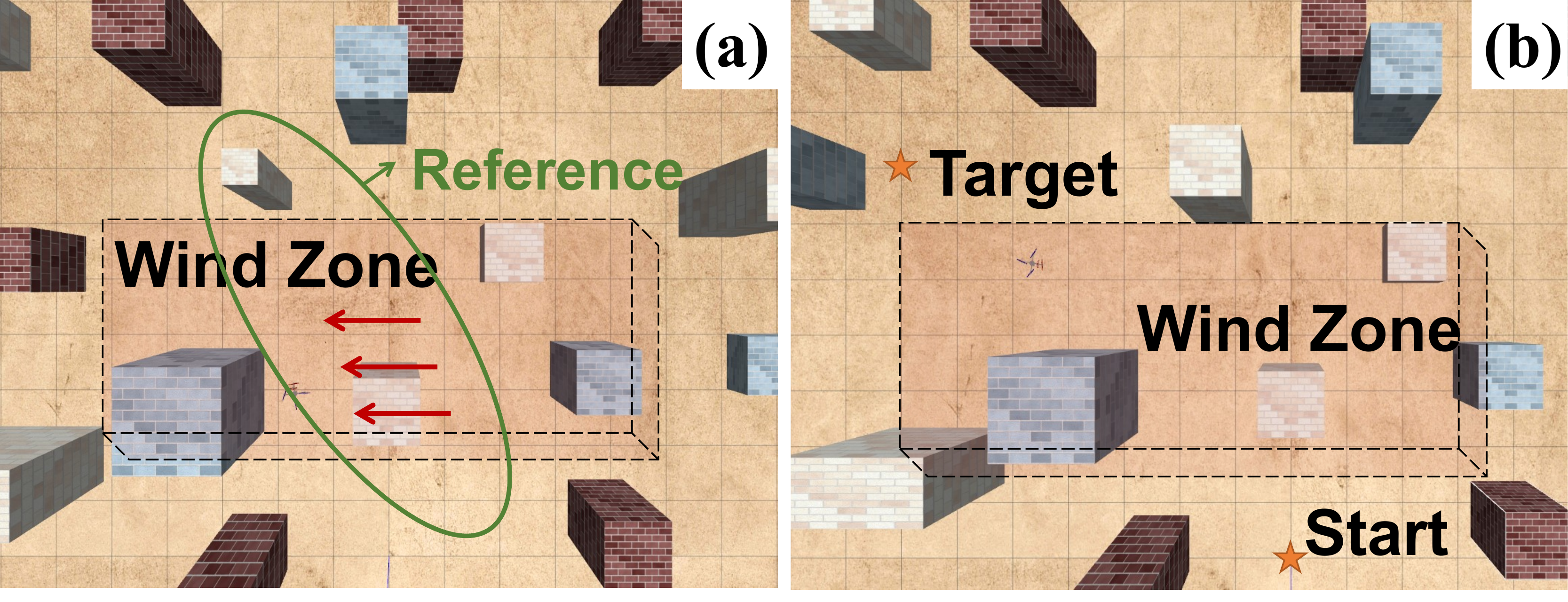}}
	\vspace{-0.3cm}
	\caption{\label{fig:fig5}The experiment settings for comparison testing. (a) The test for reference tracking. A predefined reference trajectory is provided, which is partially in the wind zone. (b) The test for local planning. With a global target, the drone aims to fly through the wind zone. }
	\end{figure}

	The RotorS MAVs simulator~\cite{Furrer2016} which includes physical engines, is employed for simulated flight tests. As shown in Fig. \ref{fig:fig5}, we separately make two scenarios of benchmark comparison. The first comparison is tracking conducted with an elliptical reference trajectory, while the second compares the whole planning framework. An equivalent command acceleration that sends to the controller is defined as $\bm{a}^{e} = [a_{x}^{e}, a_{y}^{e}, a_{z}^{e}]^{T}$ in the world frame for further evaluations. We set the maximum velocity of the quadrotor as 2.0 $\rm{m/s}$.

	The real-world tests are presented in the unknown indoor and outdoor environments. We use the autonomous quadrotor platform in~\cite{ding2020vidfusion}, with an onboard computer (i7-8550U), a stereo camera (Intel Realsense D435) for real-time mapping, a DJI N3 controller and a rotor speed measurement unit. The radius and height of our quadrotor is $r = 0.22 \ \rm{m}$ and $h = 0.13 \ \rm{m} $. The drag coefficient is identified as $k_d = 0.33$. The external force estimation setup follows the pipeline in~\cite{ding2020vidfusion}.
	\subsection{Simulation Tests}
	\begin{figure*}[!t]    
	\centering
	{\includegraphics[width=2.0\columnwidth]{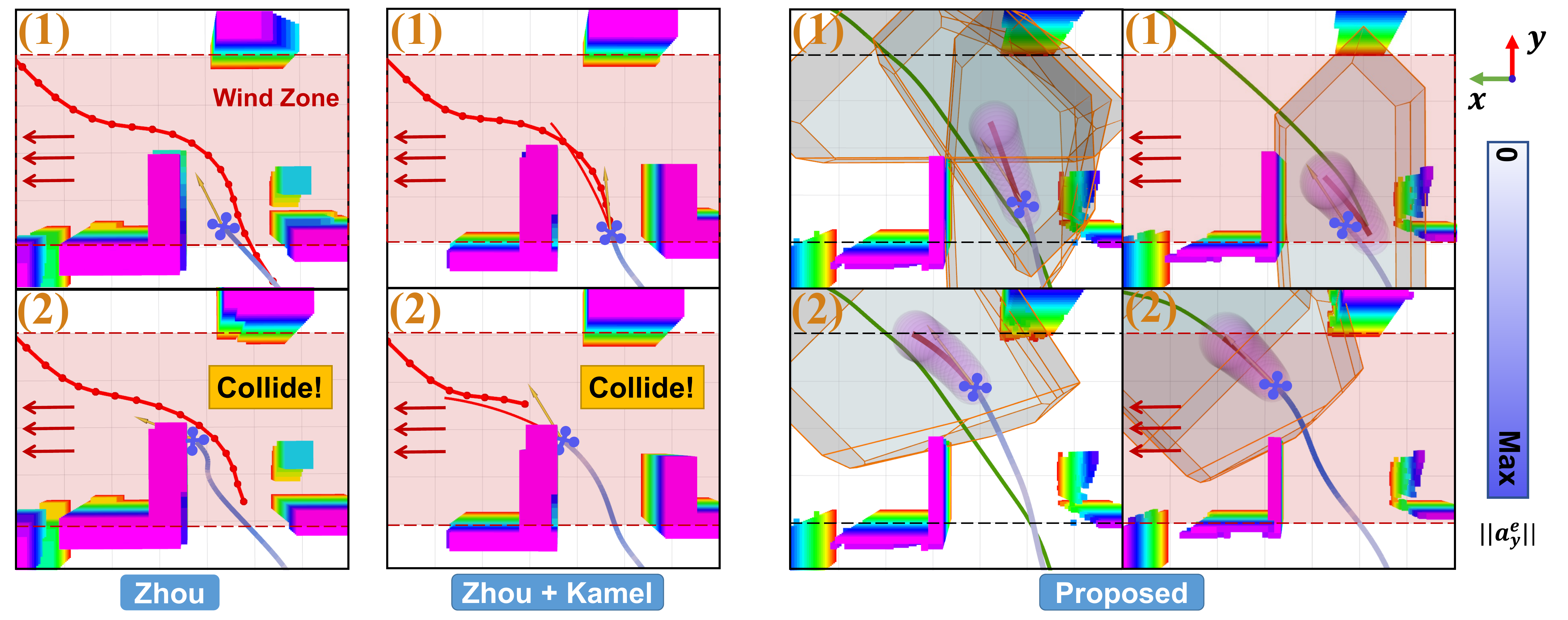}}
	\caption{\label{fig:topo}The snapshots of the simulation comparison of different methods. The sub-figure(1) shows when the quadrotor enters the wind zone, while (2) is a snapshot of when the quadrotor leaves the zone. The odometry is colored with the norm of command acceleration in the y-axis $a_{y}^{e}$. When flying through the wind zone (the red area), the quadrotor suffers a constant mass-normalized external force $\bm{\mathcal{F}}^e = [0.0, 2.0, 0.0]^T$. In Zhou's method, the quadrotor does not consider the deviation of its position to the original planning trajectory due to external force. Together with Kamel's NMPC controller, it smooths the impact of an external force, but the EKF-based estimator has some delay for the quadrotor to react. The right pictures are our method without and with a wind zone. It indicates that if the quadrotor encounters the external force, $a_{y}^{e}$ will increase for disturbance resistance.}
	\vspace{-0.3cm}
    \end{figure*}

	\subsubsection{Comparison of Reference Tracking}
	
	We firstly evaluate our external forces resilient NMPC with other MPCs~\cite{KAMEL20173463,8968126} with disturbance consideration under the same environments. Because other MPC-based methods require the predefined path and information of obstacles, for fairness to all MPCs, we provide a known global path for tracking and implement a disturbance observer for external force estimation. Method in~\cite{8968126} solves the MPC by unscented dynamic programming and needs more than 70 ms computation time for similar scale problems. Since they do not provide open source applicable code for real-time solving, we re-implement it using Forces Pro and provide available information of obstacles and wind zone for benchmark comparison. We define ``success rate'' as the rate of successful flights without collision and the control cost evaluated by squares of commanded accelerations. The simulation environment is set as Fig. \ref{fig:fig5}(a) with a global path throughout a fixed wind zone. When the quadrotor enters the range of the wind zone, it suffers from a mass-normalized external force in the y-axis to simulate different extents of the winds. 

	\begin{table}[!htbp]
		\footnotesize
		\renewcommand\arraystretch{1.15}
		\setlength{\belowcaptionskip}{-0.9cm}
		\centering
		\caption{Comparison of Reference Tracking\label{tab:test1} }
		\begin{tabular}{|c|c|c|c|c|}
			\hline 
			\tabincell{c} { $\bm{\mathcal{F}}^e $ \\ $({\rm m/s^2}) $}    &Method & \tabincell{c}{ Succ. \\ Rate} & \tabincell{c}{ Traj. \\ Time \\ (s)}  & \tabincell{c}{ Ctrl. \\ Cost \\ ($ {\rm m^2/s^3} $)}\\
			\hline 
			\multirow{3}*{$[0.0, 1.5, 0.0]^T$} &Kamel et al.~\cite{KAMEL20173463} &1.00 &15.67&18.11\\
			\cline{2-5}
			&Hoseong et al.~\cite{8968126} &1.00&15.73&19.03\\ 
			\cline{2-5}
			&Proposed  & 1.00 &14.89 & 16.34\\	
			\hline  
			\multirow{3}*{$[0.0, 2.5, 0.0]^T$} &Kamel &-&-&-\\
			\cline{2-5}
			&Hoseong &-&-&-\\ 
			\cline{2-5}
			&Proposed  & 0.60 &15.02& 27.38\\	
			\hline                    
		\end{tabular}
	\end{table}

The results are shown in Tab. \ref{tab:test1}. All MPC methods can successfully finish the flight test if the wind is within the variance range and reaction window for tracking. When the external force surpasses the bound, their methods fail to handle the disturbance while following the trajectory. 

\subsubsection{Comparison of Planning Framework}
Secondly, we compare our method against a state-of-the-art local planner by Zhou~\cite{Zhou2021EGOPlannerAE} under different external forces for real-time systems. We also test the disturbance-aware controller~\cite{KAMEL20173463} using planning trajectory from Zhou's method as a reference to assess the robustness of this controller. The scenario is as the map in Fig. \ref{fig:fig5} (2). Within the wind zone, the quadrotor is under a constant force with different values and directions.

\begin{table}[!htbp]
	\footnotesize
	\renewcommand\arraystretch{1.15}
	\setlength{\abovecaptionskip}{0.cm}
	\setlength{\belowcaptionskip}{-0.9cm}
	\centering
	\caption{Comparison of Planning Framework\label{tab:test2} }
	\begin{tabular}{|c|c|c|c|c|}
		\hline 
		\tabincell{c} { $\bm{\mathcal{F}}^e$  \\ $({\rm m/s^2}) $}    &Method & \tabincell{c}{ Succ. \\ Rate} & \tabincell{c}{ Traj. \\ Time \\ (s)}  & \tabincell{c}{ Ctrl. \\ Cost \\ ($ {\rm m^2/s^3} $)}\\
		\hline 
		\multirow{3}*{$[0.0, 0.5, 0.0]^T$}&Zhou et al.~\cite{Zhou2021EGOPlannerAE}  &0.70&7.65&14.65\\
		\cline{2-5}
		& Zhou + Kamel &0.50&10.93&19.18\\ 
		\cline{2-5}
		& Proposed  & \textbf{1.00} & 8.33 & 13.71\\	
		\hline                   
		\multirow{2}*{$[0.0, 1.0, 0.0]^T$}&Zhou&0.60&8.80&14.70\\ 
		\cline{2-5}
		& Zhou + Kamel &0.40&12.67&19.33\\ 
		\cline{2-5}
		& Proposed  & \textbf{1.00} &8.17 & 16.81\\
		\hline 	
		\multirow{2}*{$[2.0, 1.0, 0.0]^T$}&Zhou& 0.60 & 8.29 & 18.86 \\
		\cline{2-5}
		\cline{2-5}
		& Zhou + Kamel & 0.45  &9.95 & 20.01\\ 
		\cline{2-5}
		& Proposed  & \textbf{1.00} & 8.27 & 16.37\\	
		\hline 	
		\multirow{2}*{$[0.0, 2.0, 0.0]^T$}&Zhou&- & - & - \\
		\cline{2-5}
		\cline{2-5}
		& Zhou + Kamel &0.25&9.04&19.13\\ 
		\cline{2-5}
		& Proposed  & \textbf{0.90} &8.86& 12.78\\		
		\hline			
	\end{tabular}

\end{table}

	As the Tab. \ref{tab:test2} illustrates, when the external force is sufficiently small, the controller can resist such influence to a certain degree. As the external force increases, the local planner, without consideration of force disturbance, has a higher collision probability. A simulation comparison instance under mass-normalized external force equal to $ [0.0, 2.0, 0.0]^T$ is shown in Fig. \ref{fig:topo}. The local planner, without force consideration, cannot recompense the trajectory offset in the y-axis resulting from an external force. Furthermore, with a disturbance-aware observer, the robust controller cannot give real-time feedback to the planner to adjust the trajectory because the current reference is already infeasible. The EKF based disturbance observer that is applied by the controller smooths the external force. Hence, the controller performs delayed reactions to suddenly imposed force and sometimes adversely increases the collision risk. For other cases in Tab. \ref{tab:test2}, when the direction of external force is not along with the main trends of the quadrotor's motion, the safe and successful flight will cost more energy. Otherwise, the quadrotor can take advantage of the force.

	\subsection{Real-World Tests}
	\subsubsection{Indoor Flights Through Wind Zones}
	
\begin{figure}[!t]
	\begin{center}
		{\includegraphics[width=1.0\columnwidth]{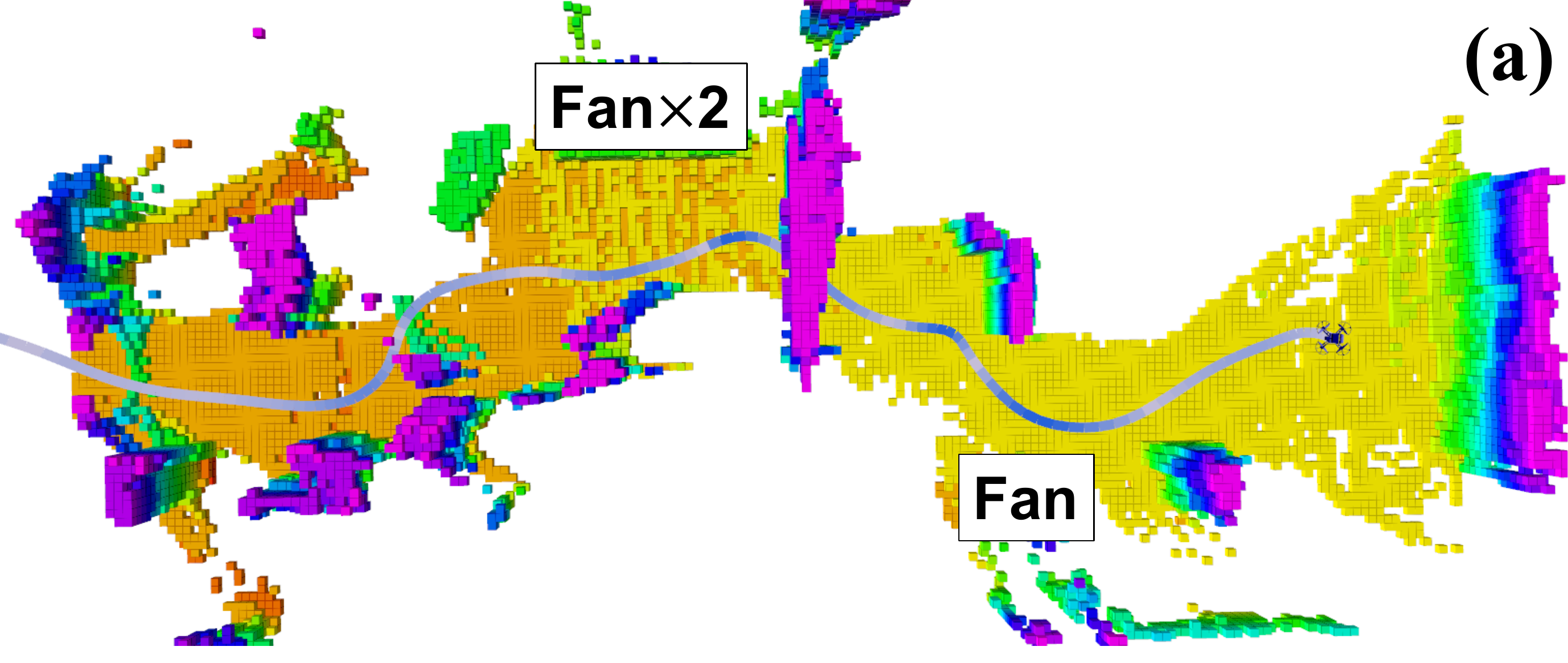}}
		{\includegraphics[width=1.01\columnwidth]{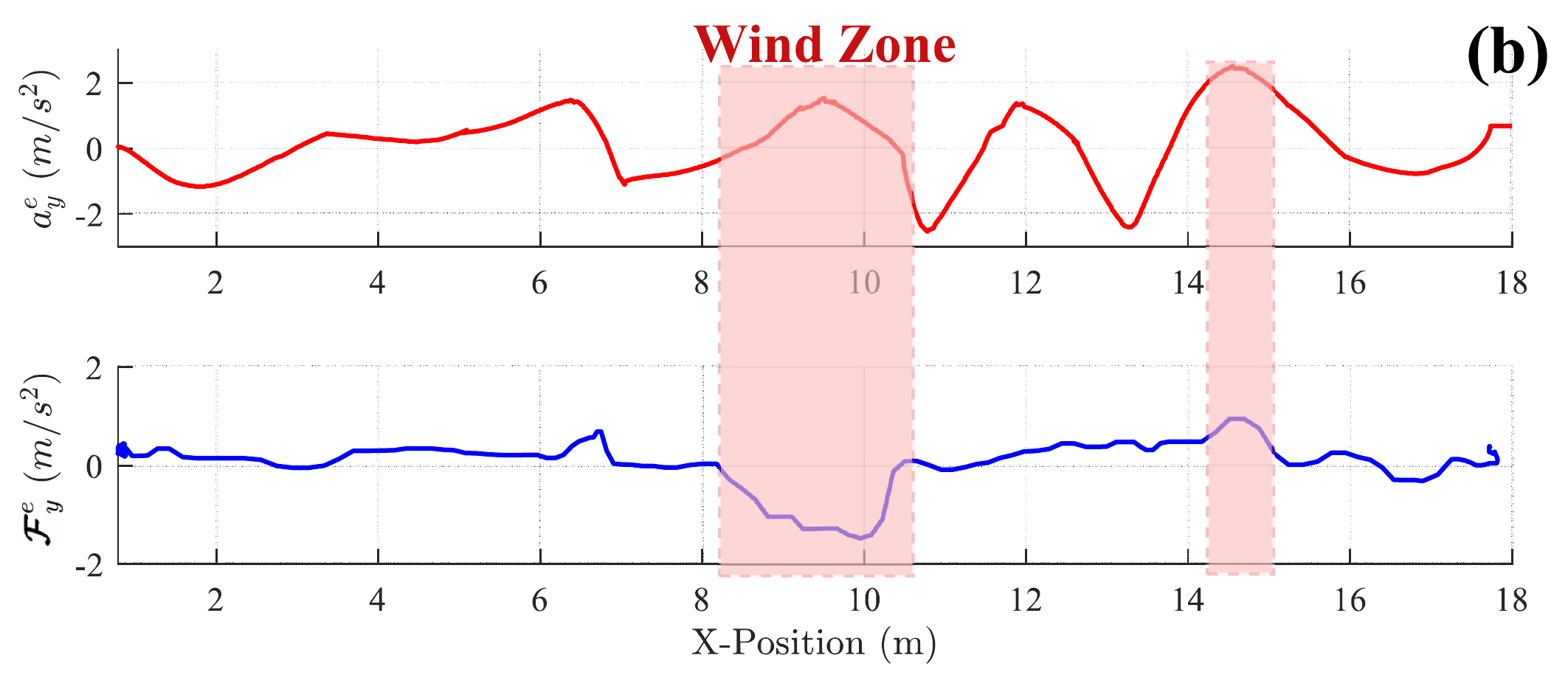}}
		\caption{\label{fig:indoor1} Real-world indoor test. (a) The overall trajectory of the indoor flight experiment, colored with the norm of command acceleration in the y-axis. (b) The variance of $\mathcal{F}_y^e$ and $a_{y}^{e}$ with x-position.}
		\vspace{-0.8cm}	
	\end{center}
\end{figure}

We present several indoor experiments in cluttered environments with strong winds in Fig. \ref{fig:fig1}. The quadrotor flies through an unknown narrow hallway fully autonomously with onboard computation and limited field of view (FOV) sensors. Several fans are set up near the possible flight paths, providing the wind zone for external disturbances. The speed of wind zones caused by fans ranges from $6.5 \ \rm{m/s}$ to $7.4 \ \rm{m/s}$, measured by an anemometer. The wind causes the maximum mass-normalized external force at around $2.5 \ \rm{m/s^2}$ to the quadrotor based on experiment data, measured by VID-Fusion~\cite{ding2020vidfusion}. The trajectory of a flight test through two different wind zones is shown in Fig. \ref{fig:indoor1}. When the quadrotor enters wind zones, the increasing estimated external force triggers its replanning, which results in the variation of acceleration in the y-axis and adjustment of starting predicted state in back-end optimization.

\subsubsection{Indoor Benchmark Flights Comparison}

\begin{figure}[!htbp]
	\vspace{-0.2cm}	
	\begin{center}
		{\includegraphics[width=1.0\columnwidth]{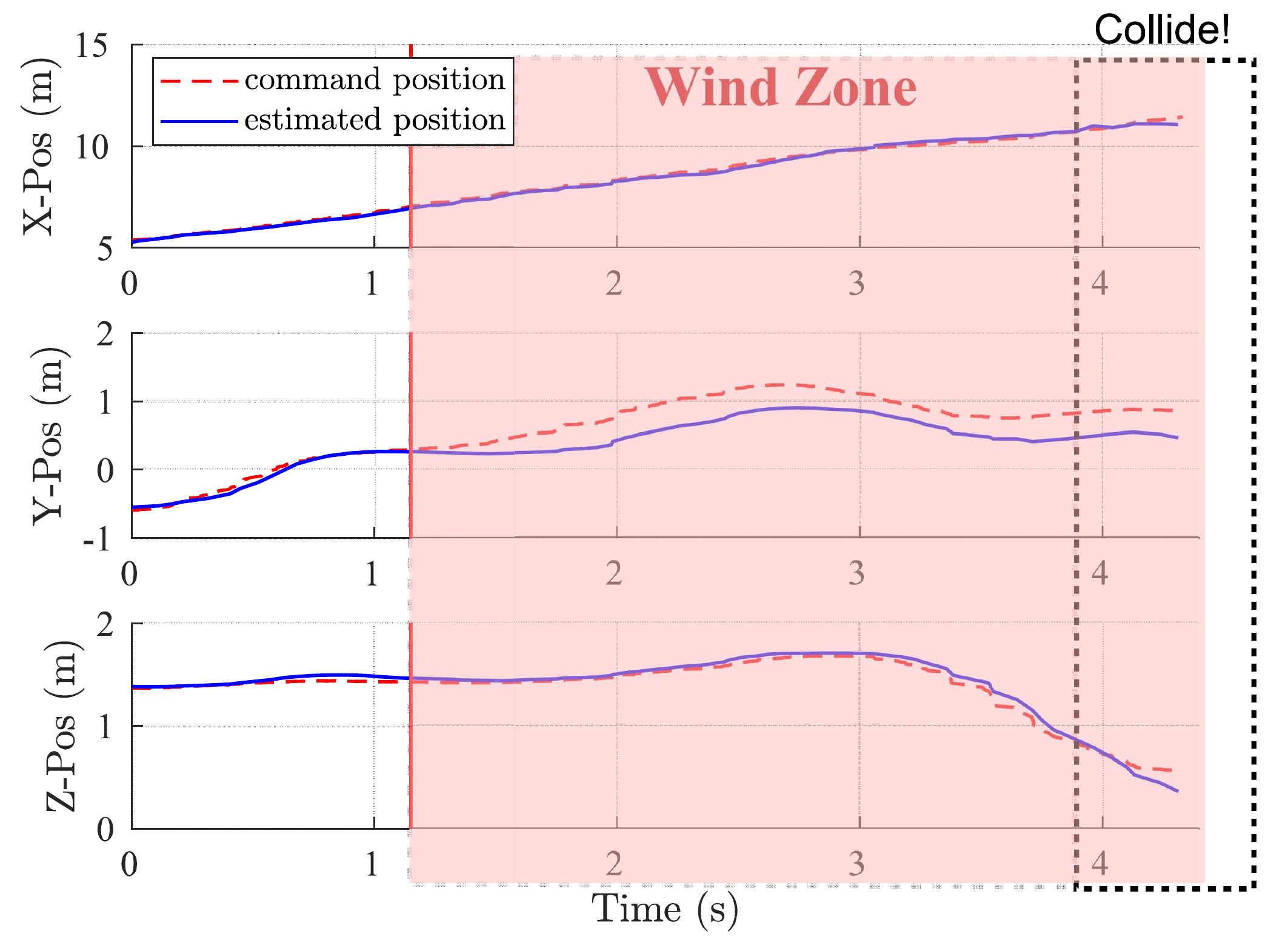}}
		\vspace{-0.5cm}	
		\caption{ \label{fig:ego_planner} Benchmark flight test for comparison. The blue curve is the estimated position by VINS, and the red doted curve is the command send by Zhou's planner over time. The red vertical marked that the quadrotor is entering a wind zone. }
		\vspace{-0.15cm}\setlength{\abovecaptionskip}{-0.5cm}
	\end{center}
	\vspace{-0.6cm}	
\end{figure}

In this section, we test Zhou's planner~\cite{Zhou2021EGOPlannerAE}  in the same indoor flight environment mentioned above, where several wind zones are presented. The results of command positions and estimated positions during one flight experiment are shown in Fig. \ref{fig:ego_planner}. After entering the wind zone, the planner fails to compensate position deviations and continues to follow an intractable trajectory, then collides with obstacles.

\subsubsection{Indoor Flights With an Unmodeled Load}

Flying with an object of unknown mass is also an essential application of delivery for quadrotors. We add a package to the quadrotor without modeling of complete dynamic model or measurement of a loaded object. A quadrotor (1.13 kg) with a loaded bottle (0.19 kg) is experimentally tested in a cluttered environment. When the quadrotor is hovering, the gravity of this load causes an equivalent mass normalized force acceleration $a_{load}=1.65 \ \rm{m/s^2}$ in the z-axis. 

\begin{figure}[!htbp]
		\vspace{-0.15cm}	
	\begin{center}
		{\includegraphics[width=1.0\columnwidth]{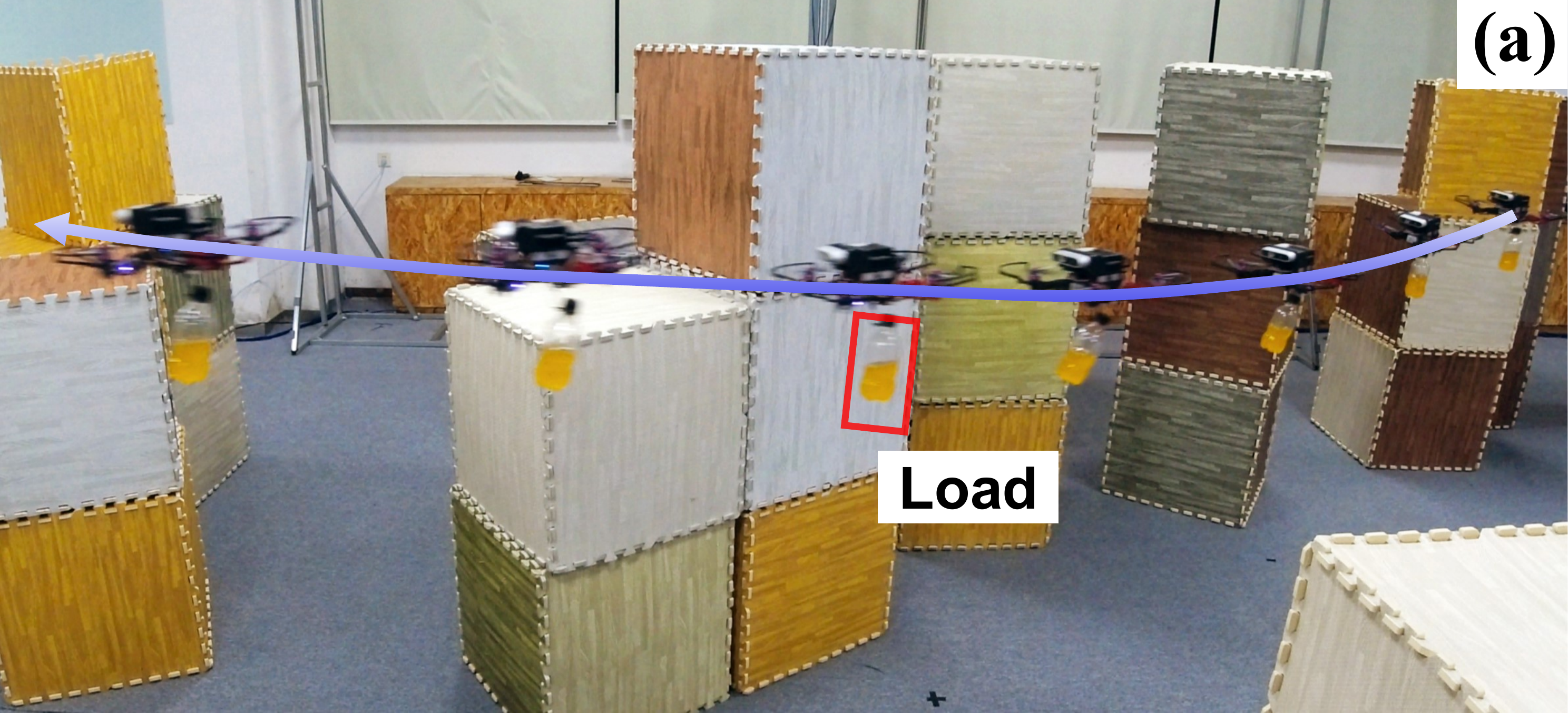}}
		{\includegraphics[width=1.0\columnwidth]{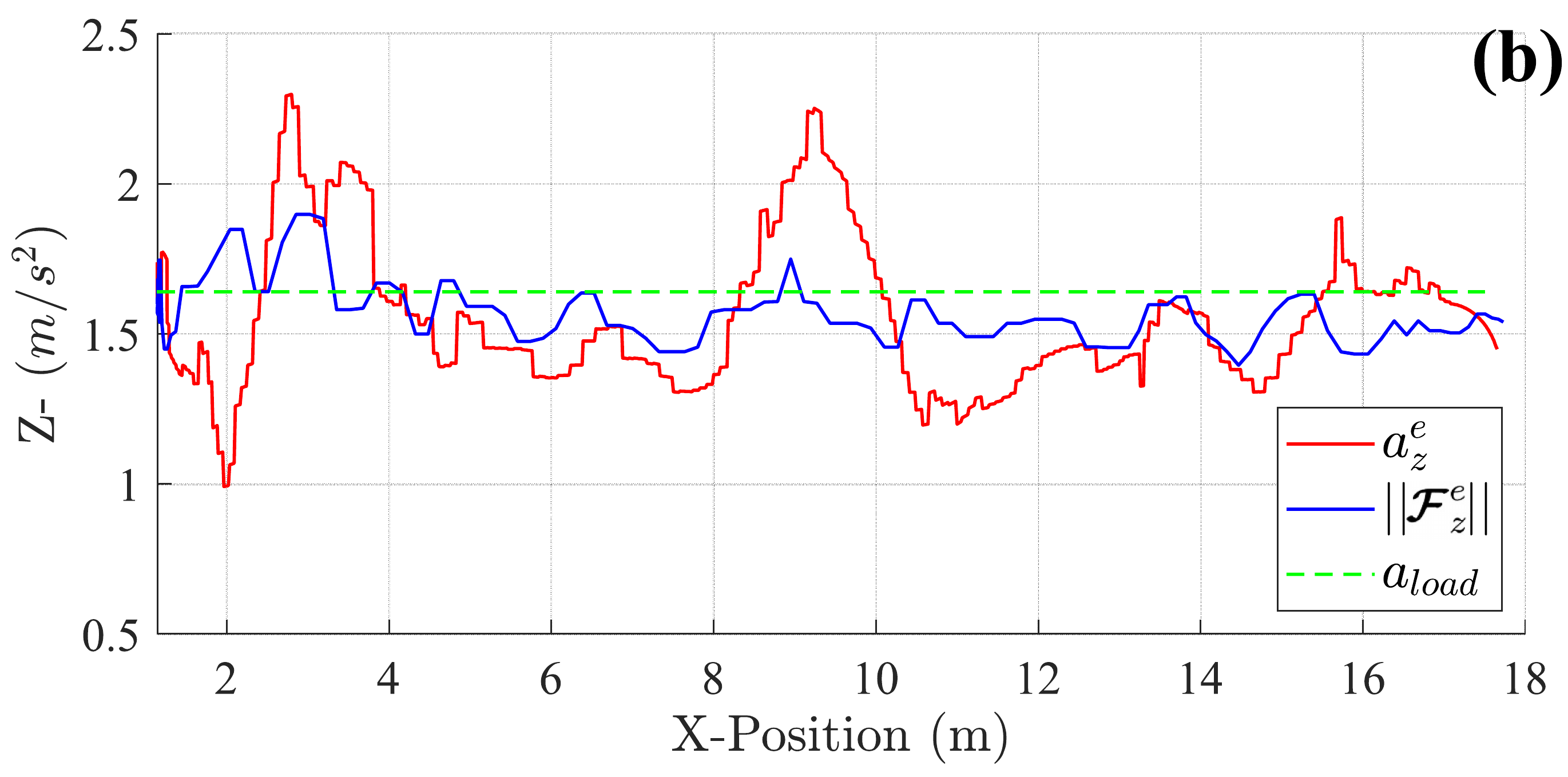}}
		\caption{\label{fig:fig9} Flight test with an unmodeled load. (a) Snapshots of indoor flight test with a loaded bottle. (b) The variance of $\mathcal{F}_z^e$ and $a_{z}^{e}$ with x-position, compared with gravity effect of the load.  }
		\vspace{-0.3cm}	
	\end{center}
\end{figure}

The snapshot is shown in Fig. \ref{fig:fig9} (1), and the estimated external force and command acceleration in the z-axis of the quadrotor is shown Fig. \ref{fig:fig9} (2). It illustrates that the planning acceleration commands sent by our planner compensate for the influence of external force, which ensures a stable trajectory in the z-direction. The value of the external force matches the gravity effect of the loaded bottle and slightly varies during the whole flight.

In our experiment, the load is connected to the quadrotor with a short string so that the pendulum effect can be neglected in path planning. If the string of a suspended load is considered, the planner needs to model its dynamics and adds obstacles avoidance of the load.

\subsubsection{Outdoor Flights Under Natural Winds}

\begin{figure}[!htbp]
	\vspace{-0.1cm}
	\centering
	{\includegraphics[width=0.99\columnwidth]{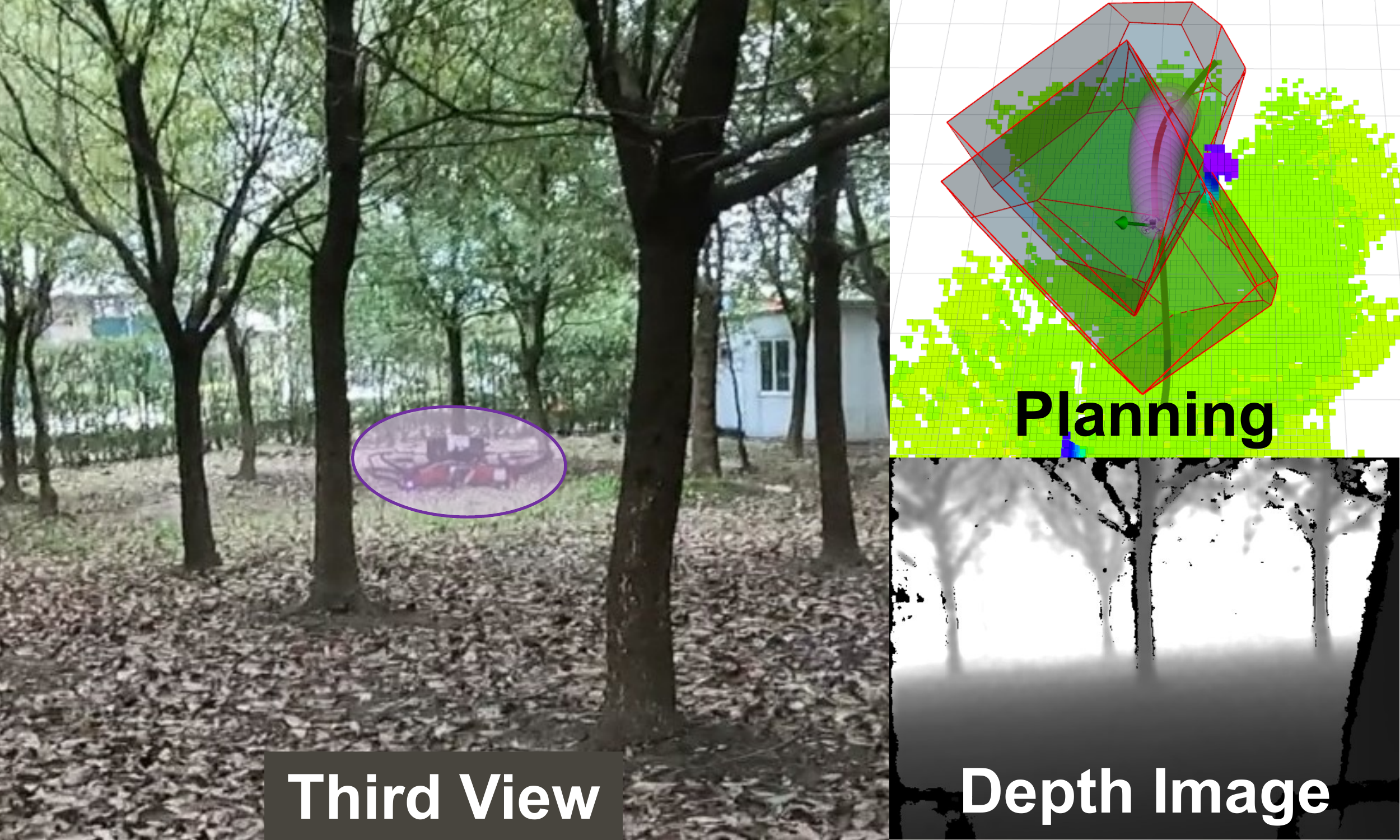}}
	\caption{\label{fig:outdoor1}The snapshot of the outdoor flight test.}
	\vspace{-0.25cm}	
\end{figure}

To validate the robustness and efficiency of the proposed planning framework under natural winds, we conduct outdoor experiments in a cluttered forest with wind speed up to $3.1\ \rm{m/s}$. Given several targets inside the forest, the drone generates a smooth trajectory to follow the targets against the natural wind gusts and other uncertain disturbances from leaves, as shown in Fig. \ref{fig:outdoor1}. It reaches a maximum velocity around $2.5\ \rm{m/s}$. The outdoor flight test shows that our proposed framework has the capability of autonomous navigation in a completely unknown and complex environment with natural disturbance.

\section{Conclusion}
\label{sec:conclusion}

This paper presents a systematic (re)planning framework for a quadrotor autonomous system in the presence of fierce external forces. We make a step forward to consistently include the external force in both the planning and tracking systems by incorporating the force in front-end path searching and back-end NMPC optimization. Using the ellipsoidal approximation of error FRSs and quadrotor's inflated geometrical shape, we can guarantee collision avoidance by constraining the safe ellipsoid boundaries into a flight corridor along the reference path. The proposed approach can achieve real-time adaptive planning without offline computation of FRSs or the pre-computed collision-free reference trajectory. The benchmark comparisons in both simulations and real-world tests illustrate the necessity and adaptability of the proposed framework with external force considerations. Furthermore, the indoor flights with different external forces and outdoor flights with natural winds validate the robustness of the proposed method under different environments. 
\vspace{-0.1cm}

    \newlength{\bibitemsep}\setlength{\bibitemsep}{.023\baselineskip}
\newlength{\bibparskip}\setlength{\bibparskip}{0pt}
\let\oldthebibliography\thebibliography
\renewcommand\thebibliography[1]{%
	\oldthebibliography{#1}%external force at around 2.5
	\setlength{\parskip}{\bibitemsep}%
	\setlength{\itemsep}{\bibparskip}%
}

\bibliography{references}
\end{document}